\documentclass{article}

\usepackage{microtype}
\usepackage{graphicx}
\usepackage{subfigure}
\usepackage{booktabs} 
\usepackage{amsmath}
\usepackage{amssymb}
\usepackage{multirow}
\usepackage{array}
\usepackage{scalerel}
\usepackage{enumitem}
\setlist[enumerate]{itemsep=0mm}

\usepackage{hyperref}
\usepackage{xcolor}

\usepackage[accepted]{icml2024}

\usepackage{amsmath}
\usepackage{amssymb}
\usepackage{mathtools}
\usepackage{amsthm}

\usepackage[capitalize,noabbrev]{cleveref}

\theoremstyle{plain}

\theoremstyle{definition}

\theoremstyle{remark}

\usepackage[textsize=tiny]{todonotes}

\icmltitlerunning{LLaGA: Large Language and Graph Assistant}

\begin{document}

\twocolumn[
\icmltitle{LLaGA: Large Language and Graph Assistant}



\icmlsetsymbol{equal}{*}




\begin{icmlauthorlist}
\icmlauthor{Runjin Chen}{UT}
\icmlauthor{Tong Zhao}{Snap}
\icmlauthor{Ajay Jaiswal}{UT}
\icmlauthor{Neil Shah}{Snap}
\icmlauthor{Zhangyang Wang}{UT}
\end{icmlauthorlist}

\icmlaffiliation{UT}{The University of Texas at Austin}
\icmlaffiliation{Snap}{Snap Inc}

\icmlcorrespondingauthor{Zhangyang Wang}{atlaswang@utexas.edu}
\icmlcorrespondingauthor{Runjin Chen}{chenrunjin@utexas.edu}


\vskip 0.3in
]



\printAffiliationsAndNotice{}  

\begin{abstract}
 Graph Neural Networks (GNNs) have empowered the advance in graph-structured data analysis. Recently, the rise of Large Language Models (LLMs) like GPT-4 has heralded a new era in deep learning. However, their application to graph data poses distinct challenges due to the inherent difficulty of translating graph structures to language.  To this end, we introduce the \textbf{L}arge \textbf{L}anguage \textbf{a}nd \textbf{G}raph \textbf{A}ssistant (\textbf{LLaGA}), an innovative model that effectively  integrates LLM capabilities to handle the complexities of graph-structured data. LLaGA retains the general-purpose nature of LLMs while adapting graph data into a format compatible with LLM input. LLaGA achieves this by reorganizing graph nodes to structure-aware sequences and then mapping these into the token embedding space through a versatile projector. LLaGA excels in versatility, generalizability and interpretability, allowing it to perform consistently well across different datasets and tasks, extend its ability to unseen datasets or tasks, and provide explanations for graphs. Our extensive experiments  across popular graph benchmarks show that LLaGA delivers outstanding performance across four datasets and three tasks using one single model, surpassing state-of-the-art graph models in both supervised and zero-shot scenarios. Our code is available at \url{https://github.com/VITA-Group/LLaGA}
 


\end{abstract}


\section{Introduction}
\label{Intro}
Graphs are omnipresent, representing a myriad of real-world data from social networks,  biological networks and recommendation systems, etc. Graph neural networks (GNNs) \cite{Kipf2017SemiSupervisedCW,defferrard2016convolutional,velivckovic2017graph}, embedded with message passing and aggregation techniques, are powerful algorithmic tools on handling complex graph structures. 
Nonetheless, a critical limitation of GNNs is their weak  multi-task handling capability. Typically trained on a single task, GNNs struggle to maintain performance when applied to multiple tasks. Self-supervised learning~\cite{jin2021automated,ju2023multitask} may offer some improvement, but they still require task-specific heads or tuning for downstream tasks.




Recently, the advent of LLMs having massive context-aware knowledge and semantic comprehension capabilities (\emph{e.g.,} LLaMa~\cite{touvron2023llama}, GPTs~\cite{achiam2023gpt}, Claude~\cite{perez2022discovering}) marks a significant advancement in AI research. A key advantage of LLMs is their ability to solve various tasks  with a single model, showcasing strong language skills and the capacity to explain provided answers.  These models have demonstrated remarkable proficiency not only in language-related tasks but also in understanding and generating visual content~\cite{liu2023visual, wang2023visionllm}. However, direct application of such models presents challenges when it comes to graph-structured data, which inherently contains rich relational and structural information. Hence, researchers ~\cite{fatemi2023talk, chen2023exploring} explored ways to translate graph structures into natural language suitable for consumption by language models. Yet, describing graphs in plain texts tends to be verbose and fails to directly represent the intrinsic characteristics of graphs, often leading to repetitive and unintuitive descriptions of nodes and edge relationships. Consequently, LLMs would perform poorly on basic graph tasks without specific adaptations~\cite{chen2023exploring}. Subsequently, InstructGLM~\cite{ye2023natural} describes graphs in language and attempts to enhance LLMs' graph-task performance by task-specific fine-tuning. However, this specialization  constrains the model's versatility, potentially limiting its effectiveness in other graph tasks or non-graph-related domains.  More recently, GraphGPT~\cite{tang2023graphgpt} has combined text descriptions with a self-supervised graph transformer to incorporate graph data into large language models (LLMs). However, the pre-trained graph transformer might not distill all relevant structural information for specific downstream tasks, leading to less satisfactory performances.
Motivated by these issues, this work poses an important \textbf{question}: 
\textit{How to develop a framework that effectively encodes structural information for graphs across various tasks and domains, enabling its comprehension by LLMs, while maintaining LLMs' general-purpose ?}

To this end, we introduce the \textbf{L}arge \textbf{L}anguage \textbf{a}nd \textbf{G}raph \textbf{A}ssistant (\textbf{LLaGA}), a novel framework that seamlessly integrates rich graph-structured data with the massive context-awareness skills and comprehension capabilities of Large Language Models. LLaGA has three impressive characteristics that distinguish LLaGA with prior works as follows:

\begin{itemize}[leftmargin=*]
\setlength\itemsep{-0.1em}
    \item \textbf{Versatility:} LLaGA adopts a simple but universally applicable method for encoding structural details in graphs, and achieves a general alignment between graph and token spaces using a single, versatile projector. 
    This projector efficiently handles various graph tasks across multiple datasets, eliminating the need for task-specific adjustments. Significantly, the performance of our versatile LLaGA framework can even exceed that of specialized task-focused graph models.
    \item \textbf{Generalizability:} Given the comprehensive alignment between graph and token spaces, LLaGA not only excels in those datasets and tasks encountered during training but also demonstrates robust generalization to previously unseen datasets and tasks without additional tuning.
    \item \textbf{Interpretability:} A key feature of LLaGA is its ability to provide detailed interpretations of node embeddings, greatly enhancing the understanding of its decision-making processes.
\end{itemize}

To achieve this, LLaGA uniquely reorganizes graph data into \textit{node sequences}, without converting structural information into potentially ambiguous natural language descriptions. These sequences are formatted with the help of novel \underline{\textbf{node-level templates}}, to reflect the structural information surrounding each central node while preserving the graph's node features. Note that this transformation is parameter-free, ensuring the preservation of the original structural integrity without necessitating further distillation. Subsequently, LLaGA translates node representations into LLMs' comprehensible token embedding space through a \textbf{\underline{versatile projector}},
which can help in mitigating the expensive computational cost of fine-tuning LLMs as well as keeping LLMs' general purpose. The projector is generally trained on multiple graph datasets across various tasks, such as node classification, link prediction, and node description. This ensures it can interpret graph data from diverse perspectives and ingest an inherent ability to handle multiple tasks (all at once), bolstering its practical utility, and potentially augmenting LLaGA's generalization capabilities across various unseen datasets and tasks. Notably, unlike traditional multi-task learning methodologies used in GNNs, LLaGA trained all tasks in a uniform Question-Answer format, eschewing the need for task-specific loss functions or heads. 
Our extensive experiments illustrate that LLaGA achieves a robust alignment between the graphs and token space of LLMs, facilitating the model's application to multiple tasks, unseen test set, and interestingly out-of-distribution datasets.  

To our best knowledge,  LLaGA is \textbf{the first single model} to preform consistently well across various graph datasets and tasks. It matches the effectiveness of specialized GNNs tailored for specific data and tasks,  while also showing strong generalizability to unseen datasets or tasks.

\section{Methodology}
\label{Method}

\begin{figure*}[t]
  \centering
  \includegraphics[width=\textwidth]{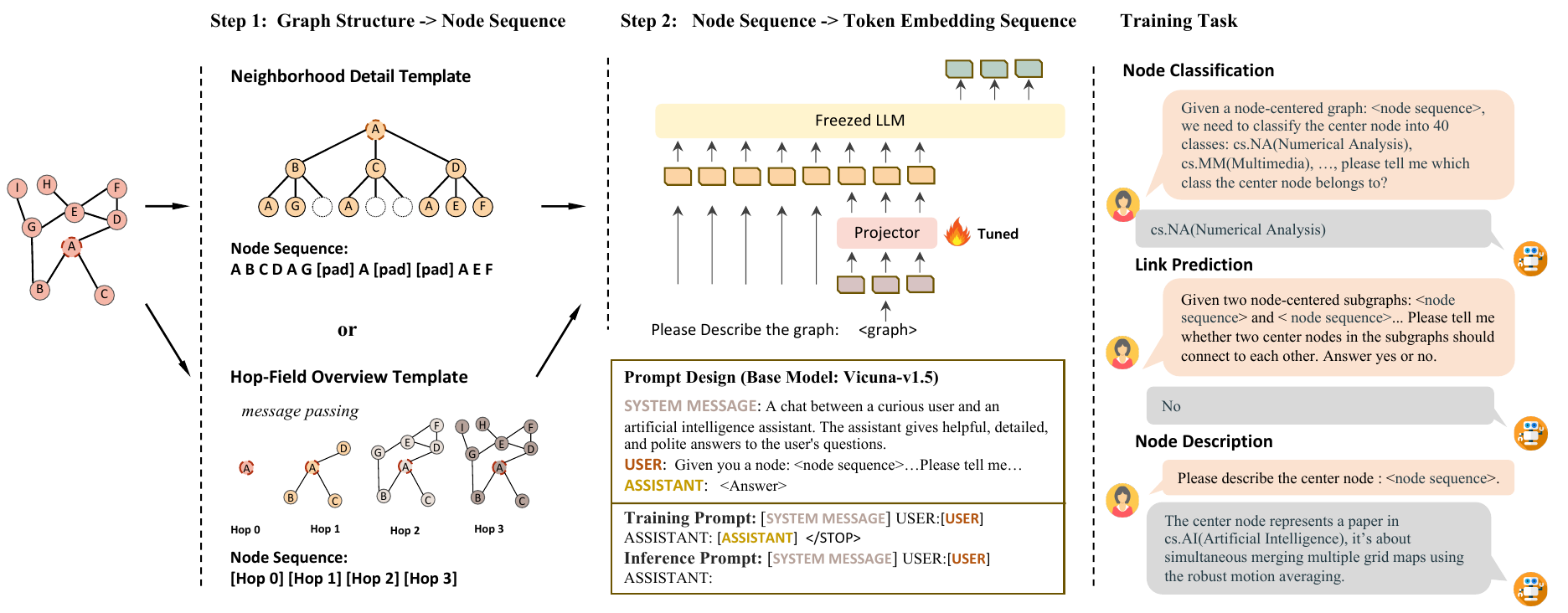}
  \vskip -0.1in
  \caption{Illustration of LLaGA framework and its prompt design paradigm.} \label{fig:main}
  \vskip -0.15in
\end{figure*}

In this section, we introduce the details of \textbf{LLaGA} framework. We start with the notation setup, followed by a detailed explanation of the method employed for translating graphs into token embedding space. Subsequently, we delve into the training process, encompassing both the design of prompts and tasks as well as the training objectives.

\subsection{Notation}
A graph is a structure that encapsulates a set of entities and the interrelationships among them. Formally, a graph is denoted as $\mathcal{G} = (\mathcal{V}, \mathcal{E}, \mathcal{X})$.
Here, $\mathcal{V}$ denotes the set of nodes (entities). The set of edges, $\mathcal{E}$, represents the connections between the nodes in $\mathcal{V}$. $\mathcal{X}$ is the attribute information corresponding to the nodes. Each node $v_i \in \mathcal{V}$ is associated with an attribute feature $x_i \in \mathcal{X}$. 
In this paper, our primary focus is on text-attributed graphs, implying that the attributes $x_i \in \mathcal{X}$ of each node are expressed in a textual format. 
Additionally, we introduce $\mathcal{N}_{v}^k$ to denote the $k^{th}$ hop neighborhood set surrounding the node $v$.

\subsection{Structure-Aware Graph Translation}
The primary objective of LLaGA (Large Language and Graph Assistant) is to translate graph inputs into a token embedding space that is comprehensible to Large Language Models. This translation enables the utilization of LLMs' inherent reasoning capabilities for graph-related tasks, without necessitating any modifications to the LLM parameters. LLaGA accomplishes this by initially reorganizing nodes in graphs into node embedding sequences. These sequences are structured according to our proposed templates and are then converted into a sequence of token embeddings using a projector. 

The first step involves converting graphs into node embedding sequences. Recognizing that the fundamental unit for graph analysis is the node, we developed two node-level templates for analysis on graphs. These templates are versatile, applicable not only to node-level tasks but also to other tasks like link prediction. Both templates are designed to encode structural information surrounding a node, offering different perspectives for analysis. The first,  the \textbf{Neighborhood Detail Template}, provides an in-depth view of the central node and its immediate surroundings. The second, the  \textbf{Hop-Field Overview Template}, offers a summarized view of a node's neighborhood, extendable to larger fields. 


\textbf{Neighborhood Detail Template}  is designed to elaborate on the detailed information of a node and its surrounding neighborhood.  Given a node $v$, we first construct a fixed-shape, sampled computational tree centered around $v$. For every hop of neighbors, we define a neighbor sample size, denoted as $n_1, n_2,...$, where $n_i$ indicates the sample size for the $i^{th}$ hop. The computational tree is built with the root node being the central node $v$. From the 1-hop neighbor set of $v$, denoted as $\mathcal{N}_{v}^1$, we randomly select $n_1$ nodes to form a new neighbor set $\widetilde{\mathcal{N}_{v}^1}$. If the size of $\mathcal{N}_{v}^1$ is smaller than $n_1$, i.e., $|\mathcal{N}_{v}^1| < n_1$, we supplement the set with placeholder nodes to reach a size of $n_1$. Therefore, the size of $\widetilde{\mathcal{N}_{v}^1}$ is consistently $n_1$, i.e., $|\widetilde{\mathcal{N}_{v}^1}| = n_1$ . The nodes in $\widetilde{\mathcal{N}_{v}^1}$ are treated as children of the root node. Subsequently, for each node in $\widetilde{\mathcal{N}_{v}^1}$, we recursively sample $n_2$ neighbors as its children. Any sets with insufficient nodes are filled with placeholder nodes. 
For any placeholder node, its children are exclusively placeholder nodes. As illustrated in upper-left of Figure~\ref{fig:main}, with the root node being $A$, we display a 2-hop neighbor structure of $A$, with the sample size of 3 for both hops. The first-order neighbors of $A$ are $\{B, C, D\}$, so they are shown in the second layer of the computational graph. Since $B$ has 2 neighbors $\{A, G\}$, we expand this set to $\{A, G, [pad]\}$, where $[pad]$ represents the placeholder node. And similarly for nodes $C$ and $D$. Ultimately, this process yields a perfect 3-ary computational tree centered around node $A$. We then perform a level-order traversal on the computational tree, transforming the comprehensive details of the central node and its neighborhood into a fixed-length node sequence. For instance, in Figure~\ref{fig:main}, the sequence representing node $A$ and its neighborhood is $A$ $B$ $C$ $D$ $A$ $G$ $[pad]$ $A$ $[pad]$ $[pad]$ $A$ $E$ $F$, where each sequence position uniquely corresponds to a relative structural position within the original graph.

Post conversion of the center node and its structural information into a node sequence, we shift to mapping them into the node embedding space. In the context of text-attributed graphs, we can utilize various off-the-shelf text encoding models $\phi$, such as SBERT~\cite{reimers2019sentence}, RoBERTa~\cite{liu2019roberta}, and SimTeG~\cite{duan2023simteg}, to encode text features. Placeholder nodes are represented by a zero vectors of the same size. We further integrate a Laplacian Embedding~\cite{dwivedi2020generalization} at each sequence position, enhancing the representation of structural information. Denoting the adjacency matrix of the computational tree by $\mathcal{A}_{tree}$, the Laplacian Embedding is defined as the eigenvectors of the Laplacian matrix of $\mathcal{A}_{tree}$:
\begin{equation}\label{sq:laplacian}
L=I-\mathcal{D}^{-\frac{1}{2}}\mathcal{A}_{tree}\mathcal{D}^{-\frac{1}{2}}=U^T\Lambda U
\end{equation}
where $\mathcal{D}$ represents the degree matrix of $\mathcal{A}_{tree}$ and $U$ symbolizes the Laplacian Embedding of the template. Notably, with a fixed sample size, the computational tree's shape remains unchanged, so the Laplacian Embedding is computed \textit{only once} for all graphs using this template. 
This Embedding is then appended to the encoded node feature to form the final node embedding. The process is outlined as follows: Let $v_1$, $v_2$, $...$, $v_n$ represent the encoded node sequence. The final node embedding $h_{v_i}$ for $v_i$ is given by
\begin{equation}\label{eq:NDemb}
h_{v_i} =
\begin{cases}
\textbf{0}\ \ ||\ \ U_i, & \text{if } v_i=[pad]; \\
\phi(x_{v_i})\ ||\ U_i, & \text{otherwise,}
\end{cases}
\end{equation}
where $||$ denotes concatenation. Subsequently, the central node and its structural information are transformed into the node embedding sequence $h_{v_1}$, $h_{v_2}$, $...$, $h_{v_n}$.

\textbf{Hop-Field Overview Template} provides a summarized view of the central node and its neighborhood. This template employs hop embeddings to characterize the node features across various neighborhood hops.  These hop embeddings are obtained through \textit{parameter-free} message passing on encoded text features. For each central node $v$, the $i^{th}$-hop embedding $h_v^i$ is calculated as follows:
\begin{equation}\label{eq:HOemb}
    h_v^i = \frac{1}{|\mathcal{N}_{v}^1|}\sum_{v'\in\mathcal{N}_{v}^1} h_{v'}^{i-1},
\end{equation}
where $h_x^0=\phi(x_{v})$. Through this calculation, $h_v^i$ potentially contains information from all neighbors in the $i^{th}$-hop neighborhood set $\mathcal{N}_{v}^i$. A sequence of hop embeddings $h_v^0$, $h_v^1$, $h_v^2$, $\ldots$ can represent the central node and its structural information. Unlike the Neighborhood Detail Template, which utilizes individual embeddings for each neighbor, the Hop-Field Overview Template summarizes each hop's neighbors with a single embedding. This approach may sacrifice some detail for the sake of a broader respective field. The choice between these templates should be based on the nature of the input data and the required level of detail.


To enhance the natural comprehension of graph inputs by Large Language Models (LLMs), it is essential to align the node embedding space with the input token space. This alignment is realized by mapping each node embedding into the token embedding space, utilizing a specifically calibrated projector, denoted as $f_{\theta}$. Mathematically, this process can be represented for a given node embedding $h_i$ as:
\begin{equation}\label{eq:projector}
e_i = f_{\theta}(h_i).
\end{equation}
Consequently, a sequence of node embeddings, $h_1$, $h_2$, ..., $h_n$, is transformed into a corresponding sequence of token embeddings, $e_1$, $e_2$, ..., $e_n$. In our framework, this transformation is facilitated by a simple MLP serving as the projector. It is important to note that the parameters $\theta$ of the projector are the only parameters subject to tuning during the training process of LLaGA.

\subsection{Alignment Tuning}

In LLaGA, we employ three key tasks on graphs – node classification, link prediction, and node description – to meticulously tune the projector. The first two tasks, node classification and link prediction, are well-established and widely recognized in the field of graph ML. Contrastingly, the node description task, which is somewhat less common in conventional graph analysis, is designed to align node embeddings with specific descriptive texts. This innovative task 
enables the provision of rich semantic interpretations of the graphs, providing a deeper insight of the logic behind graph-based predictions.The questions and answers to this task can be articulated as follows:
\textbf{Questions:} Please describe the center node: $<$node sequence$>$.
\textbf{Answers:} The center node represents a [paper / products /...], it's about [node description].  For textual-attributed graphs, the node description can be obtained from node features. By integrating these three diverse tasks into the training process, our projector develops a comprehensive and nuanced understanding of graphs and can serve as a versatile translator between node embedding and token embedding space for all those tasks. Moreover, it can explicitly generate explanations for node embeddings, enhancing interpretability. 

During training, we organize our questions and answers in a chat format. In our experiments, Vicuna~\cite{vicuna2023} serves as the primary foundational Large Language Model (LLM), so we follow the implementation strategy of Vicuna and set the system message accordingly. For details regarding the question-answer template and the training or inference input sequences, please refer to the illustrations in Figure~\ref{fig:main}. In the input processing phase, we tokenize all words in the prompt and convert them into their respective token embeddings. For the $<$node sequence$>$, we substitute this part with the projected node embeddings $e_1$, $e_2$, ..., $e_n$, maintaining their original positions. The training objective is to maximize the probability of generating the correct answer, formulated as
\begin{equation}\label{eq:max}
\underset{\theta}{\text{maximize }} p(X_{answer}|X_{graph}, X_{question}, X_{system}).
\end{equation}

\section{Experimental Results}
\label{Exp}

We conduct comprehensive experiments to validate the effectiveness of our framework across various settings, aiming to address several key research questions:
\begin{itemize}[leftmargin=*]
\setlength{\itemsep}{0pt}
\setlength{\topsep}{-0.5pt}
\item \textbf{RQ1:} How does LLaGA perform in comparison to baseline models in standard graph tasks, such as node classification and link prediction?
\item \textbf{RQ2:} How good are the interpretations generated by LLaGA for node embeddings?
\item \textbf{RQ3:} How effectively does the model transfer knowledge when adapting to new datasets or tasks in zero-shot?
\item \textbf{RQ4:} What is the contribution of our encoding templates to the overall performance?
\end{itemize}

\subsection{Setup}

\textbf{Datasets.} We train and evaluate our model on four widely-recognized graph datasets: ogbn-Arxiv~\cite{hu2020open}, ogbn-Products~\cite{hu2020open}, Pubmed, and Cora~\cite{yang2016revisiting}. These datasets span domains of citation networks and e-commerce, varying in terms of sparsity and size, ranging from small to large scales. Detailed statistics and data splitting methods are presented in Appendix~\ref{appendix:dataset}.


\textbf{Tasks.} Our model utilizes LLaGA for 3 tasks: node classification, link prediction, and graph-based node description. The targets of \textbf{\textit{node classification}} are to categorize nodes based on research topics or product characteristics. In the \textbf{\textit{link prediction}} task, we predict the existence of edges between node pairs. The \textbf{\textit{node description}} task involves generating node descriptions based on encoded node embeddings. The training ground truth is derived from classification labels and text features, structured as: \textit{The center node represents a paper/product in the [\textit{label}] domain, it's about [\textit{text feature}]}.

\textbf{Evaluation Metrics.} For evaluation metrics, we employ \textbf{\textit{Accuracy}} for both node classification and link prediction tasks,  Sbert score and Description Label Accuracy for the node description task. The \textbf{\textit{Sbert score}} measures the similarity between embeddings of the generated descriptions and the ground truth descriptions encoded by Sbert. \textbf{\textit{Description Label Accuracy}} represents the Accuracy of labels inferred from node descriptions. For LLaGA framework, a sample is considered accurate only if it precisely identifies the category's full name in its response.

\textbf{Implementation Details.}  In our model's implementation, we primarily employ Vicuna-7B-v1.5-16K~\cite{vicuna2023} as the foundational base models, and SimTeg~\cite{duan2023simteg} as default text-encoding model. Additionally, we conduct a comparative analysis of various base LLMs and embeddings in Appendix~\ref{appendix:encoding} and ~\ref{appendix:LLM}. The learning rate is consistently set to 2e-5, and the batch size is maintained at 16 for all models. We train our model for one epoch. However, to compensate for the limited data size, we replicate the training samples from the smallest dataset, Cora, three times. For the Neighborhood Detail Template, we sample two-hop neighbors around each node, setting the sample size to 10 for each hop. In the Hop-Field Overview Template, 4 hop embeddings are employed to encapsulate the structural information surrounding the central node. We denote LLaGA implementations with the Neighborhood Detail Template and Hop-Field Overview Template as \textbf{LLaGA-ND-7B} and \textbf{LLaGA-HO-7B}, respectively.

\textbf{Baselines.} In our comparative analysis, we benchmark our framework against three categories of state-of-the-art models to ensure a thorough evaluation. The first category comprises Graph Neural Networks, including GCN~\cite{kipf2016semi}, GraphSage~\cite{hamilton2017inductive}, GAT~\cite{veličković2018graph}, SGC~\cite{wu2019simplifying}, and SAGN~\cite{sun2021scalable}. The second category encompasses transformer-based graph models, NodeFormer~\cite{wu2022nodeformer}. 
The final category is represented by GPT-3.5, a leading general LLM. For the first two categories, identical text-encoding methods are employed to encode text features, ensuring a fair comparison. 
For GPT-3.5, we utilized node classification results from the survey by Chen et al.~\cite{chen2023exploring} and extended this approach to the link prediction task by employing a consistent graph-description prompt format.
In addition, we also compare with the concurrent work, GraphGPT~\cite{tang2023graphgpt}.

\subsection{Overall Performance Comparison (RQ1)}\label{sec:cmp}

\renewcommand{\arraystretch}{1.2}
\begin{table*}[t]
\caption{Performance comparison with baseline models on both node classification and link prediction under 4 settings. \textbf{\textit{Single Focus}} denotes models trained on a single task and dataset. \textbf{\textit{Task Expert}} refers to models trained exclusively on one task across all datasets, specializing in that task.\textbf{\textit{Classification Expert} }indicates models trained in both node classification and link prediction on all datasets, becoming proficient in classification tasks. \textbf{\textit{General Model}} are capable of handling classification tasks across datasets and excel in semantic tasks, such as generating interpretable descriptions for node embeddings. (\textbf{\textcolor{black}{bold}} signifies the \textbf{\textcolor{black}{best result across all methods}}, while \underline{underline} highlights the \underline{best baseline result} under this setting)  }
\vskip -0.05in
\label{tab:main}
\begin{center}
\begin{small}
\resizebox{1\textwidth}{!}{
\begin{sc}
\begin{tabular}{c|c|cccc | cccc}
\toprule
\multirow{2}{*}{Model Type} & \multirow{2}*{Model} & \multicolumn{4}{c|}{Node Classification Accuracy(\%)} & \multicolumn{4}{c}{Link Prediction Accuracy(\%)} \\
\cline{3-10}
~ & ~ & Arxiv &Products & Pubmed & Cora & Arxiv & Products & Pubmed & Cora \\
\midrule
\multirow{8}*{\shortstack{Single \\ Focus}} & GCN & 73.72 & 80.75 & 92.96 & 88.93 & 91.43 & 93.95 & \underline{90.91} & \underline{81.59} \\
~  & GraphSage & \underline{76.29} & 82.87 & 94.87 & 88.89 &  91.64 & 94.96 & 90.64 & 79.15 \\
~  & GAT & 74.06 & 83.06 & 92.33 & 88.97 & 85.99 & 93.85 & 83.96 & 80.06 \\
~  & SGC & 71.77 & 75.47 & 87.35 & 87.97 & 87.99 & 88.51 & 83.60 & 80.94 \\
~  & SAGN & 75.70 & 82.58 & \textbf{\textcolor{black}{95.17}} & \underline{89.19} & 90.62 &\underline{94.85} & 90.48 & 79.88 \\
~  & NodeFormer & 74.85 & \underline{83.72} & 94.90 & 88.23 & \underline{91.84} &  90.93&  77.69 & 77.26 \\
\cline{2-10}
~  & \textbf{LLaGA-ND-7B} & 75.98 & 84.60 & 95.03 & 88.86 & 91.24 & \textbf{\textcolor{black}{97.36}} &  \textbf{\textcolor{black}{91.41}}& 83.79 \\
~  & \textbf{LLaGA-HO-7B} & \textbf{\textcolor{black}{76.66}} & \textbf{\textcolor{black}{84.67}} & 95.03 & \textbf{\textcolor{black}{89.22}} & \textbf{\textcolor{black}{94.15}} & 95.56 & 89.18 & \textbf{\textcolor{black}{86.82}} \\
\midrule
\multirow{6}*{\shortstack{Task \\ Expert}} & GCN & 71.45 & 80.88 & 89.25 & 81.62 & \underline{88.51} & \underline{93.54} & \underline{81.01} & 78.88 \\
~  & GraphSage & \underline{72.56} & 82.50 & 94.15 & 81.99 & 87.76 & 93.49 & 76.14 & 80.74 \\
~  & GAT & 72.19 & 82.61 & 87.97 & \underline{83.58} & 82.58 &92.03 & 76.85 & 79.76 \\
~ & NodeFormer & 72.35 & \underline{82.99} & \underline{94.41} & 83.27 & 84.11 &93.42 &80.40 & \underline{81.03} \\
\cline{2-10}
~  & \textbf{LLaGA-ND-7B} & \textbf{\textcolor{black}{76.41}}& \textbf{\textcolor{black}{84.60}} & 94.78 & 88.19 & 91.20 & \textbf{\textcolor{black}{97.38}} & \textbf{\textcolor{black}{93.27}} & \textbf{\textcolor{black}{89.41}} \\
~  & \textbf{LLaGA-HO-7B} & 76.40 & 84.18 & \textbf{\textcolor{black}{95.06}} & \textbf{\textcolor{black}{89.85}} & \textbf{\textcolor{black}{94.36}} & 95.85 & 88.88 & 87.50 \\
\midrule
\multirow{6}*{\shortstack{Classification \\ Expert}} & GCN &70.95 & 80.02 &89.00 & \underline{82.77} & 87.69 & \underline{92.88} & 72.28 & 78.35 \\
~  & GraphSage &\underline{71.91} & 81.62 &\underline{91.81} & 82.44 & \underline{89.23} & 92.22 & 75.36 & 82.09 \\
~  & GAT & 70.90 & \underline{81.83} & 87.72 & 82.07 & 85.18 & 92.11 & 75.00 & 80.35 \\
~ & NodeFormer & 63.20 & 75.55 &89.50 & 69.19 & 82.33 &75.42 & \underline{78.22} & \underline{81.47} \\
\cline{2-10}
~  & \textbf{LLaGA-ND-7B} & 75.85 & \textbf{\textcolor{black}{83.58}} & \textbf{\textcolor{black}{95.06}} & 87.64 & 90.81 & \textbf{\textcolor{black}{96.56}} & \textbf{\textcolor{black}{92.36}} & 87.35 \\
~  & \textbf{LLaGA-HO-7B} & \textbf{\textcolor{black}{75.99}} & 83.32 & 94.80 & \textbf{\textcolor{black}{89.30}} & \textbf{\textcolor{black}{94.30}} & 96.05 & 88.64 & \textbf{\textcolor{black}{88.53}} \\
\midrule
\multirow{3}*{\shortstack{General \\ Model}} & GPT3.5-Turbo & \underline{55.00} & \underline{75.25} & \underline{88.00} & \underline{71.75} & \underline{63.80} & \underline{60.30} & \underline{68.70} & \underline{65.74} \\
\cline{2-10}
~  & \textbf{LLaGA-ND-7B} & 74.29 & \textbf{\textcolor{black}{82.21}} & 92.42 & \textbf{\textcolor{black}{87.82}} & 90.53 & \textbf{\textcolor{black}{96.82}} & 86.31 & 81.91 \\
~  & \textbf{LLaGA-HO-7B} & \textbf{\textcolor{black}{75.01}} & 82.07 & 94.45 & \textbf{\textcolor{black}{87.82}} & \textbf{\textcolor{black}{92.04}} & 86.80 & \textbf{\textcolor{black}{89.81}} & \textbf{\textcolor{black}{84.41}} \\

\bottomrule
\end{tabular}
\end{sc}
}
\end{small}
\end{center}
\vskip -0.15in
\end{table*}

\renewcommand{\arraystretch}{1}

\begin{table}[t]
\vskip -0.1in
\caption{Compare with Concurrent Work.}
\label{tab:Concurrent}
\vskip -0.1in
\begin{center}
\begin{small}
\resizebox{1\columnwidth}{!}{
\begin{sc}
\begin{tabular}{m{4cm}<{\centering}m{0.9cm}<{\centering}m{0.9cm}<{\centering}m{0.9cm}<{\centering}}
\toprule
\multirow{2}*{Model}  & Arxiv & Pubmed& Pubmed\\
~ & NC & NC & LP \\
\midrule
GraphGPT-Mix-7B & 64.76 & 74.16 &  58.86 \\
GraphGPT-Std-7B & 63.90 & -- & 80.26 \\
\textbf{LLaGA-ND-7B(General)} & 74.29 & 92.42 & 86.31 \\
\textbf{LLaGA-HO-7B(General) }& \textbf{75.01} & \textbf{94.45} & \textbf{89.81} \\
\bottomrule
\end{tabular}
\end{sc}
}
\end{small}
\end{center}
\vskip -0.3in
\end{table}

\begin{table*}[t]
\caption{Examples Demonstrating the Interpretability of the LLaGA Framework.}
\label{tab:example}
\vskip -0.1in
\begin{center}
\begin{small}
\resizebox{1\textwidth}{!}{
\begin{tabular}{p{17cm}}
\toprule
\textbf{Title:} Real time hand gesture detection and classification using convolutional neural networks.  \\
\textbf{Abstract:} Real-time recognition of dynamic hand gestures from video streams is a challenging task since (i) there is no indication when a gesture starts and ends in the video, (ii) performed gestures should only be recognized once, and (iii) the entire architecture should be designed considering the memory and power budget. In this work, ...
 \\
\textbf{Label: cs.CV(Computer Vision and Pattern Recognition)} \\[2pt]
\hline
\parbox[c][0.4cm][c]{0cm}{}\textbf{\textcolor{brown}{USER: }}Please describe the center node: $<$node embedding sequence$>$.\\
\textcolor{olive}{\textbf{LLaGA:}} This node represents a paper in \textbf{cs.CV(Computer Vision and Pattern Recognition)} domain, it's about learning to detect and recognize hand gestures using convolutional neural networks. \\
\midrule
\textbf{Title:} An efficient graph convolutional network technique for the travelling salesman problem.   \\
\textbf{Abstract:} This paper introduces a new learning-based approach for approximately solving the Travelling Salesman Problem on 2D Euclidean graphs. We use deep Graph Convolutional Networks to build efficient TSP graph representations and output tours in a non-autoregressive manner via highly parallelized beam search. Our approach ...  \\
\textbf{Label: cs.LG(Machine Learning)  } \\[2pt]
\hline
\parbox[c][0.4cm][c]{0cm}{}\textbf{\textcolor{brown}{USER: }}Please describe the center node: $<$node embedding sequence$>$.\\
\textcolor{olive}{\textbf{LLaGA:}} This node represents a paper in \textbf{cs.LG(Machine Learning)} domain, it's about deep graph convolutional networks for solving the traveling salesman problem. \\
\midrule
\textbf{Title:} Inverse boosting pruning trees for depression detection on twitter \\
\textbf{Abstract:} Depression is one of the most common mental health disorders, and a large number of depression people commit suicide each year... We argue that it is feasible to identify depression at an early stage by mining online social behaviours. Our approach, ... \\
\textbf{Label: cs.LG(Machine Learning)} \\[2pt]
\hline
\parbox[c][0.4cm][c]{0cm}{}\textbf{\textcolor{brown}{USER: }}Please describe the center node: $<$node embedding sequence$>$.\\
\textcolor{olive}{\textbf{LLaGA:}} This node represents a paper in \textbf{ cs.SI(Social and Information Networks)} domain,  it's about predicting suicide risk using social media data. \textcolor{red}{(Label is different from ground truth, but also reasonable)}\\

\bottomrule
\end{tabular}
}
\end{small}
\end{center}
\vskip -0.2in
\end{table*}

We compare our LLaGA model with various baselines across four distinct settings: Single Focus, Task Expert, Classification Expert, and General Model. The \textbf{\textit{Single Focus}} setting involves models trained on a single dataset for a specific task, thereby concentrating exclusively on that task. \textbf{\textit{Task Expert}} refers to models trained across all datasets but focused on a single task, enabling them to perform as specialists in that area. In the \textbf{\textit{Classification Expert}} setting, models are trained on all datasets for both node classification and link prediction tasks. The \textbf{\textit{General Model}} is trained for node classification, link prediction, and node description across all datasets, equipping the model to handle not just classification tasks but also semantic tasks like node description. The comparative results are presented in Table~\ref{tab:main}. Notably, when implementing the GNN-based or Transformer-based baselines in the Task Expert or Classification Expert settings, they were trained using a multi-task learning approach, which incorporates a shared backbone with task-specific classification heads for different datasets or tasks. In contrast, our LLaGA framework employs a single projector to handle all tasks.

\textbf{Comparision with Baselines:} Our analysis reveals three key observations. \textit{\underline{Observation 1:}  LLaGA framework demonstrates superior performance compared to baseline models across all settings, particularly in multi-task learning scenarios.} This highlights LLaGA's versatility and robust capability in addressing various graph tasks. \textit{\underline{Observation 2:} While many baseline models experience significant performance degradation in multi-task learning scenarios, LLaGA stands out by exhibiting minimal decline or even improvements in performance.} 
This reflects LLaGA's proficiency in extracting common patterns across different datasets and tasks.
Such a trait hints at the potential for developing a powerful multi-model LLM equipped with simple projectors. \textit{\underline{Observation 3:} Both the Neighborhood Detail Template and the Hop-Field Overview Template exhibit distinct advantages.} The Neighborhood Detail Template excels in tasks requiring detailed neighbor information, whereas the Hop-Field Overview Template is more effective in tasks that depend on a broader overview of neighbor information with a larger receptive field. For instance, in identifying product categories, it is illogical to classify a product as 'Video Games' based solely on many of its neighbors being 'Electronics'. A more detailed analysis, revealing numerous 'Nintendo Switch' neighbors, makes classification more accurate, as seen in the case of the ogbn-Products dataset. Conversely, for some citation graphs, an overview of a paper's neighboring categories can be more informative, making the Hop-Field Overview Template the preferable choice.

\textbf{Comparison with Concurrent Work:} We conduct a comparative analysis with our concurrent work, GraphGPT~\cite{tang2023graphgpt}. GraphGPT is a generalizable model designed for solving graph tasks using LLM. It employs a text-encoding model to extract node features and utilizes a pre-trained graph transformer for encoding structural information. In our comparison, we focus on our most robust and generalizable models, with the results detailed in Table~\ref{tab:Concurrent}, GraphGPT's results are referenced directly from its original paper. 'Mix' and 'Std' represent two categories of prompts used in GraphGPT's training process. LLaGA's most general model is trained across 12 tasks, including node classification, link prediction, and node description on datasets such as Arxiv, Products, Pubmed, and Cora. In contrast,  GraphGPT's most general model is trained on just three tasks: node classification on Arxiv and Pubmed, and link prediction on Pubmed. But our model still demonstrates superior performance on these three tasks, highlighting the efficacy of our LLaGA framework.

\subsection{Interpretation Ability Investigation (RQ2)}\label{sec:interpret}
\begin{table}[h]
\vspace{-0.1in}
\caption{Quantitative evaluation of the node description task using Sbert Score and Description Label Accuracy. The term \textbf{\textit{Base value}} refers to the mean Sbert similarity calculated between the ground truth descriptions of two randomly selected samples.}
\label{tab:nd}
\vspace{-0.1in}
\begin{center}
\begin{small}
\resizebox{1\columnwidth}{!}{
\begin{sc}
\begin{tabular}{m{1.4cm}<{\centering}|m{2.3cm}<{\centering}|m{0.9cm}<{\centering}m{0.9cm}<{\centering}m{0.9cm}<{\centering}}
\toprule
Dataset  & Model & Base value & Sbert score & ACC\\
\midrule
\multirow{2}*{Arxiv} & LLaGA-ND-7B & \multirow{2}*{0.2231} & 0.6023 & 74.64 \\
 ~ & LLaGA-HO-7B & ~ & 0.6228 & 75.49 \\
\midrule
\multirow{2}*{Products} & LLaGA-ND-7B &\multirow{2}*{0.1513} & 0.4952 & 83.18 \\
 ~ & LLaGA-HO-7B & ~ & 0.5193 &  84.60\\
\midrule
 \multirow{2}*{Pubmed} & LLaGA-ND-7B & \multirow{2}*{0.4869} & 0.6847 & 92.27 \\
 ~ & LLaGA-HO-7B &  ~ & 0.6934 & 94.27 \\
 \midrule
 \multirow{2}*{Cora} & LLaGA-ND-7B & \multirow{2}*{0.3221} & 0.6465 & 86.72 \\
 ~ & LLaGA-HO-7B & ~ & 0.6545 & 86.90 \\
\bottomrule
\end{tabular}
\end{sc}
}
\end{small}
\end{center}
\vskip -0.1in
\end{table}
As previously stated, our LLaGA framework excels in providing comprehensive interpretations of node embeddings. We initially assess LLaGA's performance in the node description task using several quantitative metrics, with results presented in Table~\ref{tab:nd}. The \textbf{\textit{Sbert Score}} indicates the semantic similarity between the ground truth and LLaGA-generated text, measured using Sbert. We also include a \textbf{\textit{Base value}} for your reference, representing the average similarity across two randomly chosen samples. Notably, LLaGA's Sbert score significantly exceeds this base value, demonstrating its effectiveness in generating meaningful and relevant descriptions for node embeddings. Furthermore, the high accuracy in extracting labels from these descriptions corroborates the precision of the generated content.

To further illustrate this, Table~\ref{tab:example} showcases sample descriptions. These examples indicate the high quality of text produced by LLaGA. Even in some instances where LLaGA's label predictions diverge from the ground truth, its results are found to be reasonable and LLaGA effectively utilizes its generated text to substantiate these plausible interpretations.

\subsection{Zero-Shot Ability Investigation (RQ3)}\label{sec:zero}
\begin{table}[h]
\vspace{-0.1in}
\caption{Zero-Shot on Link Prediction}
\label{tab:zero}
\vspace{-0.1in}
\begin{center}
\begin{small}
\resizebox{1.05\columnwidth}{!}{
\begin{sc}
\begin{tabular}{ccc}
\toprule
Train $\rightarrow$ Test  & Model & Accuracy\\
\midrule
\multirow{5}*{\shortstack{Arxiv+Pubmed \\ $\downarrow$ \\ Cora}} &GCN &58.97 \\
~  & GraphSage & 67.68 \\
~  & GraphGPT-7B & 50.74 \\
~  & \textbf{LLaGA-ND-7B} & \textbf{86.47} \\
~  & \textbf{LLaGA-HO-7B} & \textbf{87.35} \\
\midrule
\multirow{5}*{\shortstack{Arxiv+Pubmed+Cora \\ $\downarrow$ \\ Products}} &GCN & 56.73  \\
~  & GraphSage & 58.92 \\
~  & GraphGPT-7B & 50.74 \\
~  & \textbf{LLaGA-ND-7B} & \textbf{92.65} \\
~  & \textbf{LLaGA-HO-7B} & \textbf{92.99} \\
\bottomrule
\end{tabular}
\end{sc}
}
\end{small}
\end{center}
\vskip -0.2in
\end{table}

In this section, we illustrate the generalization capabilities of LLaGA, concentrating on the task of link prediction within a zero-shot setting. For analysis of generalization capabilities in  node classification tasks, please refer to Appendix~\ref{appendix:nodezero}.

Zero-shot learning entails training a model on certain datasets and subsequently evaluating it on unseen datasets or tasks. 
This approach is instrumental in assessing a model's proficiency in transferring knowledge. 
In our study, we examine LLaGA's zero-shot performance in both in-domain and out-of-domain transfer scenarios. For in-domain transfer, the model is trained on the Arxiv and Pubmed datasets and evaluated on the Cora dataset. All three datasets comprise citation graphs. Conversely, for out-of-domain transfer, training is conducted on the Arxiv, Pubmed, and Cora datasets, with the evaluation on the Products dataset. Here, while the training datasets are citation graphs, the test set consists of e-commerce graphs. The results, as presented in Table~\ref{tab:zero}, reveal that our model exhibits robust zero-shot capabilities in both scenarios. This indicates that LLaGA can effectively discern and leverage similar patterns across datasets, adeptly transferring knowledge not only to analogous data but also to datasets that markedly differ in domain.

\subsection{Templates Ablation Study (RQ4)}\label{sec:ablation}
\begin{table}[t]
\caption{Templates Ablation Study.}
\label{tab:ablation}
\vspace{-0.1in}
\begin{center}
\begin{small}
\resizebox{1\columnwidth}{!}{
\begin{sc}
\begin{tabular}{m{0.8cm}<{\centering}m{1.2cm}<{\centering}m{0.9cm}<{\centering}m{1.2cm}<{\centering}m{0.9cm}<{\centering}m{0.9cm}<{\centering}}
\toprule
Task  & Template & Arxiv & Products & Pubmed & Cora\\
\midrule
\multirow{3}*{NC} &None &73.92& 80.45 & 94.60& 84.50 \\
~   &ND &75.85& 83.58 & 95.06 & 87.64 \\
~   &HO &75.99& 83.32 & 94.80 & 89.30 \\
\midrule
\multirow{3}*{LP} &None &89.98& 91.73 & 78.19 & 83.97 \\
~   &ND &90.81& 96.56 & 92.36 & 87.35 \\
~   &HO &94.30& 96.05 & 88.64 & 88.53 \\
\bottomrule
\end{tabular}
\end{sc}
}
\end{small}
\end{center}
\vskip -0.3in
\end{table}
We conduct an ablation study to investigate the individual contributions of our encoding templates. For this, we train a new model in a classification expert setting, but without using a template. This model solely relies on the embedding of the center node for prediction, rather than a node embedding sequence that encapsulates structural information surrounding the center node. The results are presented in Table~\ref{tab:ablation}. It is evident that both the Neighborhood Detail Template and the Hop-Field Overview Template significantly enhance performance compared to the model without a template. This is particularly noticeable in the link prediction task, a task that heavily relies on structural information. All these findings underscore the effectiveness of our templates in encoding the structural information of nodes.

\section{Related Work}
\label{Related}
\subsection{Graph Neural Networks}

GNNs have long been at the forefront of graph machine learning. 
They are designed to transform input nodes into compact vector representations, suitable for subsequent classification tasks when paired with a classification head. A common strategy among many GNNs~\cite{kipf2016semi,veličković2018graph,xu2018powerful,Gao2018LargeScaleLG,chiang2019cluster,You2020L2GCNLA,chen2018fastgcn,thekumparampil2018attention}, involves a layer-wise message-passing mechanism. This approach enables nodes to progressively aggregate and process information from their immediate neighbors, thereby embedding the nodes into lower-dimensional spaces. Concurrently, a growing body of research~\cite{yun2019graph, ying2021transformers, wu2022nodeformer,chen2022nagphormer}, has been exploring the integration of transformer-based encoders for graph data analysis, opening new avenues for enhancing GNN capabilities. However, a significant limitation of traditional graph models is their poor task generalization capability. GNNs are usually trained on a single classification task. When applied to a variety of datasets or downstream tasks, these models often fail to perform consistently well across all tasks with one single model~\cite{ju2023multitask}. 

\subsection{Self-Supervised Learning for GNNs}
Recent advancements have employed self-supervised learning strategies on GNNs to bolster their generalization performance. These methods encompass developing specialized pretext tasks for graph structures, such as mutual information maximization~\cite{velivckovic2019deep,hassani2020contrastive}, whitening decorrelation~\cite{zhang2021canonical}, and generative reconstruction~\cite{hou2022graphmae}. Moreover, investigations into integrating multi-task learning with self-supervised learning paradigms have been conducted, offering novel insights into enhancing model generalization ability~\cite{ju2023multitask}. However, these methods still require task-specific classification heads and tuning  for every downstream task, after obtaining a general embedding from the graph encoder.
\subsection{Large Language Models for Graphs}

Recent studies have explored integrating Large Language Models (LLMs) with GNNs, leveraging LLMs' extensive knowledge for graph data enhancement. Research has focused on augmenting GNNs with LLMs to enrich graph textual attributes~\cite{ye2023natural, chen2023label, tang2023graphgpt, guo2023gpt4graph, he2023harnessing, huang2023can}, though these approaches largely depend on GNNs for predictions, potentially limiting their scope. Alternatively, efforts to linguistically represent graphs for direct LLM processing encountered difficulties in effectively translating structures into natural language, often yielding suboptimal results~ \cite{huang2023can,guo2023gpt4graph}. While fine-tuning LLMs for graphs can improve performance on specific tasks, it may also limit the LLMs' versatility. GraphGPT~\cite{tang2023graphgpt} sought to address these challenges by using a pretrained graph transformer for encoding graph structures for LLMs, though finding a universally applicable graph model proved difficult. Our contribution diverges by introducing a novel encoding method that translates graph data into sequences directly compatible with LLMs, avoiding the need for intermediary models. This method shows superior versatility and generalizability across a range of tasks, even in zero-shot scenarios, outperforming traditional graph models.

\section{Conclusion}\label{sec:conclusion}
This paper introduces LLaGA, an innovative framework that effectively integrates Large Language Models (LLMs) into the graph domain while preserving their proficiency in other tasks. Instead of using complex language for describing structure information, LLaGA employs templates to transform graph structure into sequences, and then maps node embeddings to token embedding spaces using a tuned projector. This projector establishes a comprehensive alignment between texts and graphs, enabling the use of LLMs for fundamental graph tasks like node classification and link prediction across various datasets. And it can be further generalized to unseen datasets or tasks without any adaption. Additionally, it facilitates the generation of textual explanations for node embeddings. Through extensive evaluations in different settings, our method has demonstrated its effectiveness in both supervised and zero-shot graph learning scenarios. 

\section{Impact Statements}\label{sec:impact}
Our research introduces LLaGA, a novel framework that seamlessly blends the capabilities of Large Language Models (LLMs) with graph structures, enhancing the versatility of LLMs to perform fundamental graph tasks. The broader impact of LLaGA extends to numerous fields where graph data is pivotal, including but not limited to, bioinformatics, social network analysis, and knowledge graphs. As we push the boundaries of Machine Learning and AI, we recognize the importance of monitoring for unintended consequences, such as the perpetuation of biases or misuse of predictive insights. To this end, we encourage continued ethical evaluation and the development of guidelines to ensure that the applications of LLaGA contribute constructively to society. This work aspires to be a stepping stone towards more sophisticated, equitable, and transparent AI systems that respect the intricate structure of data across various domains.

\nocite{langley00}

\bibliography{main}

\begin{thebibliography}{45}
\providecommand{\natexlab}[1]{#1}
\providecommand{\url}[1]{\texttt{#1}}
\expandafter\ifx\csname urlstyle\endcsname\relax
  \providecommand{\doi}[1]{doi: #1}\else
  \providecommand{\doi}{doi: \begingroup \urlstyle{rm}\Url}\fi

\bibitem[Achiam et~al.(2023)Achiam, Adler, Agarwal, Ahmad, Akkaya, Aleman, Almeida, Altenschmidt, Altman, Anadkat, et~al.]{achiam2023gpt}
Achiam, J., Adler, S., Agarwal, S., Ahmad, L., Akkaya, I., Aleman, F.~L., Almeida, D., Altenschmidt, J., Altman, S., Anadkat, S., et~al.
\newblock Gpt-4 technical report.
\newblock \emph{arXiv preprint arXiv:2303.08774}, 2023.

\bibitem[Chen et~al.(2018)Chen, Ma, and Xiao]{chen2018fastgcn}
Chen, J., Ma, T., and Xiao, C.
\newblock Fastgcn: fast learning with graph convolutional networks via importance sampling.
\newblock \emph{arXiv preprint arXiv:1801.10247}, 2018.

\bibitem[Chen et~al.(2022)Chen, Gao, Li, and He]{chen2022nagphormer}
Chen, J., Gao, K., Li, G., and He, K.
\newblock Nagphormer: A tokenized graph transformer for node classification in large graphs.
\newblock In \emph{The Eleventh International Conference on Learning Representations}, 2022.

\bibitem[Chen et~al.(2023{\natexlab{a}})Chen, Mao, Li, Jin, Wen, Wei, Wang, Yin, Fan, Liu, et~al.]{chen2023exploring}
Chen, Z., Mao, H., Li, H., Jin, W., Wen, H., Wei, X., Wang, S., Yin, D., Fan, W., Liu, H., et~al.
\newblock Exploring the potential of large language models (llms) in learning on graphs.
\newblock \emph{arXiv preprint arXiv:2307.03393}, 2023{\natexlab{a}}.

\bibitem[Chen et~al.(2023{\natexlab{b}})Chen, Mao, Wen, Han, Jin, Zhang, Liu, and Tang]{chen2023label}
Chen, Z., Mao, H., Wen, H., Han, H., Jin, W., Zhang, H., Liu, H., and Tang, J.
\newblock Label-free node classification on graphs with large language models (llms).
\newblock \emph{arXiv preprint arXiv:2310.04668}, 2023{\natexlab{b}}.

\bibitem[Chiang et~al.(2019)Chiang, Liu, Si, Li, Bengio, and Hsieh]{chiang2019cluster}
Chiang, W.-L., Liu, X., Si, S., Li, Y., Bengio, S., and Hsieh, C.-J.
\newblock Cluster-gcn: An efficient algorithm for training deep and large graph convolutional networks.
\newblock In \emph{Proceedings of the 25th ACM SIGKDD International Conference on Knowledge Discovery \& Data Mining}, pp.\  257--266, 2019.

\bibitem[Chiang et~al.(2023)Chiang, Li, Lin, Sheng, Wu, Zhang, Zheng, Zhuang, Zhuang, Gonzalez, Stoica, and Xing]{vicuna2023}
Chiang, W.-L., Li, Z., Lin, Z., Sheng, Y., Wu, Z., Zhang, H., Zheng, L., Zhuang, S., Zhuang, Y., Gonzalez, J.~E., Stoica, I., and Xing, E.~P.
\newblock Vicuna: An open-source chatbot impressing gpt-4 with 90\%* chatgpt quality, March 2023.
\newblock URL \url{https://lmsys.org/blog/2023-03-30-vicuna/}.

\bibitem[Defferrard et~al.(2016)Defferrard, Bresson, and Vandergheynst]{defferrard2016convolutional}
Defferrard, M., Bresson, X., and Vandergheynst, P.
\newblock Convolutional neural networks on graphs with fast localized spectral filtering.
\newblock \emph{Advances in neural information processing systems}, 29, 2016.

\bibitem[Duan et~al.(2023)Duan, Liu, Chua, Yan, Ooi, Xie, and He]{duan2023simteg}
Duan, K., Liu, Q., Chua, T.-S., Yan, S., Ooi, W.~T., Xie, Q., and He, J.
\newblock Simteg: A frustratingly simple approach improves textual graph learning.
\newblock \emph{arXiv preprint arXiv:2308.02565}, 2023.

\bibitem[Dwivedi \& Bresson(2020)Dwivedi and Bresson]{dwivedi2020generalization}
Dwivedi, V.~P. and Bresson, X.
\newblock A generalization of transformer networks to graphs.
\newblock \emph{arXiv preprint arXiv:2012.09699}, 2020.

\bibitem[Fatemi et~al.(2023)Fatemi, Halcrow, and Perozzi]{fatemi2023talk}
Fatemi, B., Halcrow, J., and Perozzi, B.
\newblock Talk like a graph: Encoding graphs for large language models.
\newblock \emph{arXiv preprint arXiv:2310.04560}, 2023.

\bibitem[Gao et~al.(2018)Gao, Wang, and Ji]{Gao2018LargeScaleLG}
Gao, H., Wang, Z., and Ji, S.
\newblock Large-scale learnable graph convolutional networks.
\newblock \emph{Proceedings of the 24th ACM SIGKDD International Conference on Knowledge Discovery \& Data Mining}, 2018.

\bibitem[Guo et~al.(2023)Guo, Du, and Liu]{guo2023gpt4graph}
Guo, J., Du, L., and Liu, H.
\newblock Gpt4graph: Can large language models understand graph structured data? an empirical evaluation and benchmarking.
\newblock \emph{arXiv preprint arXiv:2305.15066}, 2023.

\bibitem[Hamilton et~al.(2017)Hamilton, Ying, and Leskovec]{hamilton2017inductive}
Hamilton, W., Ying, Z., and Leskovec, J.
\newblock Inductive representation learning on large graphs.
\newblock \emph{Advances in neural information processing systems}, 30, 2017.

\bibitem[Hassani \& Khasahmadi(2020)Hassani and Khasahmadi]{hassani2020contrastive}
Hassani, K. and Khasahmadi, A.~H.
\newblock Contrastive multi-view representation learning on graphs.
\newblock In \emph{International conference on machine learning}, pp.\  4116--4126. PMLR, 2020.

\bibitem[He et~al.(2023)He, Bresson, Laurent, Perold, LeCun, and Hooi]{he2023harnessing}
He, X., Bresson, X., Laurent, T., Perold, A., LeCun, Y., and Hooi, B.
\newblock Harnessing explanations: Llm-to-lm interpreter for enhanced text-attributed graph representation learning.
\newblock \emph{arXiv preprint arXiv:2305.19523}, 2023.

\bibitem[Hou et~al.(2022)Hou, Liu, Cen, Dong, Yang, Wang, and Tang]{hou2022graphmae}
Hou, Z., Liu, X., Cen, Y., Dong, Y., Yang, H., Wang, C., and Tang, J.
\newblock Graphmae: Self-supervised masked graph autoencoders.
\newblock In \emph{Proceedings of the 28th ACM SIGKDD Conference on Knowledge Discovery and Data Mining}, pp.\  594--604, 2022.

\bibitem[Hu et~al.(2020)Hu, Fey, Zitnik, Dong, Ren, Liu, Catasta, and Leskovec]{hu2020open}
Hu, W., Fey, M., Zitnik, M., Dong, Y., Ren, H., Liu, B., Catasta, M., and Leskovec, J.
\newblock Open graph benchmark: Datasets for machine learning on graphs.
\newblock \emph{Advances in neural information processing systems}, 33:\penalty0 22118--22133, 2020.

\bibitem[Huang et~al.(2023)Huang, Zhang, Mei, and Ma]{huang2023can}
Huang, J., Zhang, X., Mei, Q., and Ma, J.
\newblock Can llms effectively leverage graph structural information: when and why.
\newblock \emph{arXiv preprint arXiv:2309.16595}, 2023.

\bibitem[Jin et~al.(2021)Jin, Liu, Zhao, Ma, Shah, and Tang]{jin2021automated}
Jin, W., Liu, X., Zhao, X., Ma, Y., Shah, N., and Tang, J.
\newblock Automated self-supervised learning for graphs.
\newblock \emph{arXiv preprint arXiv:2106.05470}, 2021.

\bibitem[Ju et~al.(2023)Ju, Zhao, Wen, Yu, Shah, Ye, and Zhang]{ju2023multitask}
Ju, M., Zhao, T., Wen, Q., Yu, W., Shah, N., Ye, Y., and Zhang, C.
\newblock Multi-task self-supervised graph neural networks enable stronger task generalization.
\newblock In \emph{The Eleventh International Conference on Learning Representations}, 2023.
\newblock URL \url{https://openreview.net/forum?id=1tHAZRqftM}.

\bibitem[Kipf \& Welling(2017)Kipf and Welling]{Kipf2017SemiSupervisedCW}
Kipf, T. and Welling, M.
\newblock Semi-supervised classification with graph convolutional networks.
\newblock \emph{ArXiv}, abs/1609.02907, 2017.

\bibitem[Kipf \& Welling(2016)Kipf and Welling]{kipf2016semi}
Kipf, T.~N. and Welling, M.
\newblock Semi-supervised classification with graph convolutional networks.
\newblock In \emph{International Conference on Learning Representations}, 2016.

\bibitem[Langley(2000)]{langley00}
Langley, P.
\newblock Crafting papers on machine learning.
\newblock In Langley, P. (ed.), \emph{Proceedings of the 17th International Conference on Machine Learning (ICML 2000)}, pp.\  1207--1216, Stanford, CA, 2000. Morgan Kaufmann.

\bibitem[Liu et~al.(2023)Liu, Li, Wu, and Lee]{liu2023visual}
Liu, H., Li, C., Wu, Q., and Lee, Y.~J.
\newblock Visual instruction tuning.
\newblock \emph{arXiv preprint arXiv:2304.08485}, 2023.

\bibitem[Liu et~al.(2019)Liu, Ott, Goyal, Du, Joshi, Chen, Levy, Lewis, Zettlemoyer, and Stoyanov]{liu2019roberta}
Liu, Y., Ott, M., Goyal, N., Du, J., Joshi, M., Chen, D., Levy, O., Lewis, M., Zettlemoyer, L., and Stoyanov, V.
\newblock Roberta: A robustly optimized bert pretraining approach.
\newblock \emph{arXiv preprint arXiv:1907.11692}, 2019.

\bibitem[Perez et~al.(2022)Perez, Ringer, Lukošiūtė, Nguyen, et~al.]{perez2022discovering}
Perez, E., Ringer, S., Lukošiūtė, K., Nguyen, K., et~al.
\newblock Discovering language model behaviors with model-written evaluations, 2022.
\newblock URL \url{https://arxiv.org/abs/2212.09251}.

\bibitem[Reimers \& Gurevych(2019)Reimers and Gurevych]{reimers2019sentence}
Reimers, N. and Gurevych, I.
\newblock Sentence-bert: Sentence embeddings using siamese bert-networks.
\newblock In \emph{Proceedings of the 2019 Conference on Empirical Methods in Natural Language Processing and the 9th International Joint Conference on Natural Language Processing (EMNLP-IJCNLP)}, pp.\  3982--3992, 2019.

\bibitem[Sun et~al.(2021)Sun, Gu, and Hu]{sun2021scalable}
Sun, C., Gu, H., and Hu, J.
\newblock Scalable and adaptive graph neural networks with self-label-enhanced training.
\newblock \emph{arXiv preprint arXiv:2104.09376}, 2021.

\bibitem[Tang et~al.(2023)Tang, Yang, Wei, Shi, Su, Cheng, Yin, and Huang]{tang2023graphgpt}
Tang, J., Yang, Y., Wei, W., Shi, L., Su, L., Cheng, S., Yin, D., and Huang, C.
\newblock Graphgpt: Graph instruction tuning for large language models.
\newblock \emph{arXiv preprint arXiv:2310.13023}, 2023.

\bibitem[Thekumparampil et~al.(2018)Thekumparampil, Wang, Oh, and Li]{thekumparampil2018attention}
Thekumparampil, K.~K., Wang, C., Oh, S., and Li, L.-J.
\newblock Attention-based graph neural network for semi-supervised learning.
\newblock \emph{arXiv preprint arXiv:1803.03735}, 2018.

\bibitem[Touvron et~al.(2023)Touvron, Lavril, Izacard, Martinet, Lachaux, Lacroix, Rozi{\`e}re, Goyal, Hambro, Azhar, et~al.]{touvron2023llama}
Touvron, H., Lavril, T., Izacard, G., Martinet, X., Lachaux, M.-A., Lacroix, T., Rozi{\`e}re, B., Goyal, N., Hambro, E., Azhar, F., et~al.
\newblock Llama: Open and efficient foundation language models.
\newblock \emph{arXiv preprint arXiv:2302.13971}, 2023.

\bibitem[Veli{\v{c}}kovi{\'c} et~al.(2017)Veli{\v{c}}kovi{\'c}, Cucurull, Casanova, Romero, Lio, and Bengio]{velivckovic2017graph}
Veli{\v{c}}kovi{\'c}, P., Cucurull, G., Casanova, A., Romero, A., Lio, P., and Bengio, Y.
\newblock Graph attention networks.
\newblock \emph{arXiv preprint arXiv:1710.10903}, 2017.

\bibitem[Veli{\v{c}}kovi{\'c} et~al.(2019)Veli{\v{c}}kovi{\'c}, Fedus, Hamilton, Li{\`o}, Bengio, and Hjelm]{velivckovic2019deep}
Veli{\v{c}}kovi{\'c}, P., Fedus, W., Hamilton, W.~L., Li{\`o}, P., Bengio, Y., and Hjelm, R.~D.
\newblock Deep graph infomax.
\newblock 2019.

\bibitem[Veličković et~al.(2018)Veličković, Cucurull, Casanova, Romero, Liò, and Bengio]{veličković2018graph}
Veličković, P., Cucurull, G., Casanova, A., Romero, A., Liò, P., and Bengio, Y.
\newblock Graph attention networks.
\newblock In \emph{International Conference on Learning Representations}, 2018.
\newblock URL \url{https://openreview.net/forum?id=rJXMpikCZ}.

\bibitem[Wang et~al.(2023)Wang, Chen, Chen, Wu, Zhu, Zeng, Luo, Lu, Zhou, Qiao, et~al.]{wang2023visionllm}
Wang, W., Chen, Z., Chen, X., Wu, J., Zhu, X., Zeng, G., Luo, P., Lu, T., Zhou, J., Qiao, Y., et~al.
\newblock Visionllm: Large language model is also an open-ended decoder for vision-centric tasks.
\newblock \emph{arXiv preprint arXiv:2305.11175}, 2023.

\bibitem[Wu et~al.(2019)Wu, Souza, Zhang, Fifty, Yu, and Weinberger]{wu2019simplifying}
Wu, F., Souza, A., Zhang, T., Fifty, C., Yu, T., and Weinberger, K.
\newblock Simplifying graph convolutional networks.
\newblock In \emph{International conference on machine learning}, pp.\  6861--6871. PMLR, 2019.

\bibitem[Wu et~al.(2022)Wu, Zhao, Li, Wipf, and Yan]{wu2022nodeformer}
Wu, Q., Zhao, W., Li, Z., Wipf, D.~P., and Yan, J.
\newblock Nodeformer: A scalable graph structure learning transformer for node classification.
\newblock \emph{Advances in Neural Information Processing Systems}, 35:\penalty0 27387--27401, 2022.

\bibitem[Xu et~al.(2018)Xu, Hu, Leskovec, and Jegelka]{xu2018powerful}
Xu, K., Hu, W., Leskovec, J., and Jegelka, S.
\newblock How powerful are graph neural networks?
\newblock In \emph{International Conference on Learning Representations}, 2018.

\bibitem[Yang et~al.(2016)Yang, Cohen, and Salakhudinov]{yang2016revisiting}
Yang, Z., Cohen, W., and Salakhudinov, R.
\newblock Revisiting semi-supervised learning with graph embeddings.
\newblock In \emph{International conference on machine learning}, pp.\  40--48. PMLR, 2016.

\bibitem[Ye et~al.(2023)Ye, Zhang, Wang, Xu, and Zhang]{ye2023natural}
Ye, R., Zhang, C., Wang, R., Xu, S., and Zhang, Y.
\newblock Natural language is all a graph needs.
\newblock \emph{arXiv preprint arXiv:2308.07134}, 2023.

\bibitem[Ying et~al.(2021)Ying, Cai, Luo, Zheng, Ke, He, Shen, and Liu]{ying2021transformers}
Ying, C., Cai, T., Luo, S., Zheng, S., Ke, G., He, D., Shen, Y., and Liu, T.-Y.
\newblock Do transformers really perform badly for graph representation?
\newblock \emph{Advances in Neural Information Processing Systems}, 34:\penalty0 28877--28888, 2021.

\bibitem[You et~al.(2020)You, Chen, Wang, and Shen]{You2020L2GCNLA}
You, Y., Chen, T., Wang, Z., and Shen, Y.
\newblock L2-gcn: Layer-wise and learned efficient training of graph convolutional networks.
\newblock \emph{2020 IEEE/CVF Conference on Computer Vision and Pattern Recognition (CVPR)}, pp.\  2124--2132, 2020.

\bibitem[Yun et~al.(2019)Yun, Jeong, Kim, Kang, and Kim]{yun2019graph}
Yun, S., Jeong, M., Kim, R., Kang, J., and Kim, H.~J.
\newblock Graph transformer networks.
\newblock \emph{Advances in neural information processing systems}, 32, 2019.

\bibitem[Zhang et~al.(2021)Zhang, Wu, Yan, Wipf, and Yu]{zhang2021canonical}
Zhang, H., Wu, Q., Yan, J., Wipf, D., and Yu, P.~S.
\newblock From canonical correlation analysis to self-supervised graph neural networks.
\newblock \emph{Advances in Neural Information Processing Systems}, 34:\penalty0 76--89, 2021.

\end{thebibliography}
\bibliographystyle{icml2024}

\newpage
\appendix
\onecolumn


\section{Dataset Statistics}\label{appendix:dataset}
\begin{table}[h]
\caption{Dataset Statistics}
\label{tab:dataset}
\vskip 0.1in
\begin{center}
\begin{small}
\begin{tabular}{ccccc}
\toprule
Dataset & Domain & \#Node & \#Edge &Sparsity(\textpertenthousand) \\
\midrule
Cora & citation & 2708 & 5429 & 14.8065 \\
Pubmed & citation & 19717 & 44338 & 2.2810 \\
Arxiv & citation & 169343 & 1166243 & 0.8134 \\
Products & e-commerce & 2449029	 & 61859140 & 0.2063 \\
\bottomrule
\end{tabular}
\end{small}
\end{center}
\end{table}
In citation graphs (ogbn-Arxiv, Pubmed, Cora), each node represents a paper, where the title and abstract serve as node features, and edges denote co-citations. For ogbn-Products, nodes represent Amazon products, featuring item descriptions as node features, with edges indicating co-purchases.

\textbf{Data Split.} For node-level tasks, we adhere to the standard train/validation/test splits~\cite{hu2020open} for each dataset: 6:2:3 for Arxiv, 8:2:90 for Products, and 6:2:2 for both Pubmed and Cora. For link prediction, we randomly select node pairs from the node-level training set for training and from the node-level test set for testing, ensuring the edge-level training sets are equal in size to the node-level training sets.

\section{Zero-Shot Ability on Node Classification}\label{appendix:nodezero}
\renewcommand{\arraystretch}{1.2}
 \renewcommand{\arraystretch}{1.3}
\begin{table}[h]
\vspace{-0.1in}
\caption{Zero-Shot on Node Classification}
\vspace{0.1in}
\label{tab:zeronc}
\begin{center}
\begin{small}
\begin{sc}
\begin{tabular}{c|c|c|cc}
\toprule
Train $\rightarrow$ Test  & Prompt Type & Model & Accuracy(\%)\\
\midrule
\multirow{4}*{\shortstack{Arxiv+Pubmed  $\rightarrow$  Cora \\ \\ (Test task: 7 categories)}} & \multirow{2}*{Only Node Embedding} & GraphGPT-7B & 8.30\\
~  & ~ &  LLaGA-7B & \textbf{34.69} \\
\cline{2-4}
~ & \multirow{2}*{Node Embedding+Text Attributes} & GraphGPT-7B & 44.65 \\
~  & ~ &  LLaGA-7B & \textbf{59.59} \\
\midrule
\multirow{4}*{\shortstack{Arxiv+Pubmed+Cora  $\rightarrow$  Products \\ \\ (Test task: 47 categories)}} & \multirow{2}*{Only Node Embedding} & GraphGPT-7B & 1.40 \\
~  & ~ &  LLaGA-7B & \textbf{13.89} \\
\cline{2-4}
~ & \multirow{2}*{Node Embedding+Text Attributes} & GraphGPT-7B & 18.84 \\
~  & ~ &  LLaGA-7B & \textbf{43.79} \\
\bottomrule
\end{tabular}
\end{sc}
\end{small}
\end{center}
\vskip -0.1in
\end{table}

To explore the generalization capabilities of LLaGA, we also employed zero-shot learning for node classification tasks. Unlike link prediction tasks, applying zero-shot learning to node classification presents greater challenges due to the distinct label sets and the varied knowledge requirements across tasks. However, a universal aspect potentially transferable across all node classification tasks is the alignment between the graph and the semantic token space. To this end, we trained models on node description tasks from certain datasets to establish a generalized alignment between the graph structure and the token space, subsequently testing this alignment on node classification tasks using different datasets. Furthermore, we assessed LLaGA's zero-shot performance in both in-domain and out-of-domain transfer scenarios. In the in-domain scenario, training was performed on citation graphs (Arxiv + Pubmed), with testing conducted also on citation graphs (Cora).  However, the out-of-domain scenario involved training on citation graphs (Arxiv + Pubmed + Cora), with testing on the e-commerce graphs (Products). Since traditional GNNs depend on task-specific classification heads and new classification tasks may vary in label sets, they are unable to conduct zero-shot learning on node classification tasks. Our comparison was limited to llm-based baselines, specifically GraphGPT.

Our evaluation encompasses two kinds of prompts. In the first prompt, the model is only supplied with node embedding sequences, containing both attribute and structural information of the central node. The second prompt enhances this by also incorporating the textual attributes of the central node to assist the model. As detailed in Table~\ref{tab:zeronc}, our findings reveal that LLaGA consistently outperforms GraphGPT across all settings. This superiority is attributed to LLaGA's comprehensive alignment between the graph space and the token space. Moreover, the inclusion of the central node's textual attributes appears to offer some advantages in zero-shot scenarios. However, prompts based solely on node sequence embeddings show potential for application to graphs whose node attributes are challenging to describe textually, such as non-textual graphs.

\section{Flexibility with Text Encoding Methods}\label{appendix:encoding}

\renewcommand{\arraystretch}{1.2}
\begin{table*}[h]
\vskip -0.02in
\caption{LLaGA Trained with SBert and Roberta Embedding.}
\label{tab:emb}
\vskip -0.02in
\begin{center}
\begin{small}
\resizebox{1\textwidth}{!}{
\begin{sc}
\begin{tabular}{m{2cm}<{\centering}|m{2.5cm}<{\centering}|m{1cm}<{\centering}m{1.3cm}<{\centering}m{1cm}<{\centering}m{1cm}<{\centering}|m{1cm}<{\centering}m{1.3cm}<{\centering}m{1cm}<{\centering}m{1cm}<{\centering}}
\toprule
\multirow{2}*{Embedding}  & \multirow{2}*{Model} & \multicolumn{4}{c|}{Node Classification Accuracy} & \multicolumn{4}{c}{Link Prediction Accuracy}\\
\cline{3-10}
~ & ~ & Arxiv & Products & Pubmed & Cora & Arxiv & Products & Pubmed & Cora \\
\midrule
\multirow{3}*{Sbert} &GCN &66.00& 77.41 & 82.04& 79.70  &91.38& 94.91 & 84.31 & 83.15 \\
~   &GraphSage &66.79&76.00 & 82.74 & 80.66 &88.18&94.23 & 78.38 &83.62 \\
~   &\textbf{LLaGA} &\textbf{74.46}& \textbf{80.70} & \textbf{90.04} &\textbf{88.56} &\textbf{93.68}& \textbf{96.84} & \textbf{91.39} &\textbf{87.79} \\
\midrule
\multirow{3}*{Roberta} &GCN &66.51& 77.74 & 80.04& 79.30 &91.01&94.66 & 80.94 & 81.03 \\
~   &GraphSage &68.14&76.73 & 81.27 & 82.29 &88.80&94.11&74.31 &82.88 \\
~   &\textbf{LLaGA} &\textbf{74.19}& \textbf{81.13} & \textbf{89.78} &\textbf{88.19} &\textbf{93.52}& \textbf{96.79} & \textbf{89.96} & \textbf{85.15} \\

\bottomrule
\end{tabular}
\end{sc}
}
\end{small}
\end{center}
\end{table*}
LLaGA demonstrates flexibility in its text encoding methods for node attributes. In our initial experiments, we employed SimTeG~\cite{duan2023simteg} as the primary encoding model. This section also explores the use of SBERT~\cite{reimers2019sentence} and RoBERTa~\cite{liu2019roberta} as alternative encoding methods. The outcomes of these trials are shown in Table~\ref{tab:emb}. All models, including baselines, underwent training in a classification expert setting. For LLaGA, we utilized the Hop-Field Overview Template for structure encoding. Notably, LLaGA consistently surpassed other leading GNNs in performance, regardless of the chosen encoding model.

\section{Integration with Various LLMs}\label{appendix:LLM}
\renewcommand{\arraystretch}{1.2}
\begin{table*}[h]
\caption{Integration with Various LLMs}
\label{tab:base}
\begin{center}
\begin{small}
\resizebox{1\textwidth}{!}{
\begin{sc}
\begin{tabular}{m{3cm}<{\centering}|m{1.3cm}<{\centering}m{1.3cm}<{\centering}m{1.3cm}<{\centering}m{1.3cm}<{\centering}|m{1.3cm}<{\centering}m{1.3cm}<{\centering}m{1.3cm}<{\centering}m{1.3cm}<{\centering}}
\toprule
\multirow{2}*{Base Model} & \multicolumn{4}{c|}{Node Classification Accuracy} & \multicolumn{4}{c}{Link Prediction Accuracy}\\
\cline{2-9}
 ~ & Arxiv & Products & Pubmed & Cora & Arxiv & Products & Pubmed & Cora \\
\midrule
Vicuna-7B  & 75.99 & 83.32 & 94.80 & 89.30 & 94.30 & 96.05 & 88.64 & 88.53 \\
LLAMA2-7B & 76.26 & 84.21 & 94.83 & 86.53 & 94.15 & 96.03 & 89.39 & 85.44 \\
OPT-2.7B & 75.66 & 83.01 & 95.01 & 88.38 & 93.36 & 92.83 & 86.92 & 89.41\\

\bottomrule
\end{tabular}
\end{sc}
}
\end{small}
\end{center}
\end{table*}

LLaGA also demonstrates flexibility with various Base Large Language Models (LLMs). In our primary experiments, Vicuna-7B served as the foundational model. This section details the substitution of LLaGA's base LLM with alternative models, including LLaMA2-7B and OPT-2.7B. The outcomes of these replacements are presented in Table~\ref{tab:base}. For structural encoding, we employ the Hop-Field Overview Template. And models are trained in classification setting. It is evident that LLaGA consistently yields favorable results irrespective of the base LLM, showcasing its effectiveness even with comparatively lighter models such as OPT-2.7B.

\section{Experiment Variance}\label{appendix:variance}
\begin{table}[h]
\vspace{-0.1in}
\caption{Variance Information on Cora and Pubmed Dataset}
\vspace{0.1in}
\label{tab:var}
\begin{center}
\begin{small}
\begin{sc}
\begin{tabular}{c|c|c|cc}
\toprule
Setting  & Dataset & Model & NC(\%) & LP(\%)\\
\midrule
\multirow{4}*{Single Focus} & \multirow{2}*{Cora} &LLaGA-ND-7B & 88.86$\pm$0.78 & 83.79$\pm$1.26\\
~  & ~ & LLaGA-HO-7B &  89.22$\pm$0.46 &  86.82$\pm$0.88 \\
\cline{2-5}
~ & \multirow{2}*{Pubmed} &LLaGA-ND-7B & 95.03$\pm$0.12 & 91.41$\pm$0.21\\
~  & ~ & LLaGA-HO-7B &  95.03$\pm$0.07 &  89.18$\pm$0.34 \\
\bottomrule
\end{tabular}
\end{sc}
\end{small}
\end{center}
\vskip -0.1in
\end{table}

We perform training and inference five times on relatively small datasets, with the variance information detailed in Table~\ref{tab:var}.

\end{document}